\documentclass[a4paper,fleqn]{cas-sc}

\usepackage{times}
\usepackage{epsfig}
\usepackage{graphicx}
\usepackage{amsmath}
\usepackage{amssymb}

\usepackage{graphicx}
\usepackage{multicol}
\usepackage{subfig}
\usepackage{amsmath}
\usepackage{amsfonts}
\usepackage{romannum}
\usepackage{caption}
\usepackage{tabularx}
\usepackage{tabulary}
\usepackage{array}
\usepackage{tabu}
\usepackage{xcolor}
\usepackage{verbatim}
\usepackage{siunitx}
\usepackage{mathtools}

\usepackage{floatrow}
\usepackage{kotex}
\usepackage[export]{adjustbox}

\usepackage{floatrow}   

\usepackage{graphicx,xcolor} 
\usepackage[framemethod=tikz]{mdframed}

\newcolumntype{L}[1]{>{\raggedright\arraybackslash}m{#1}} %p
\newcolumntype{C}[1]{>{\centering\arraybackslash}m{#1}}
\newcolumntype{R}[1]{>{\raggedleft\arraybackslash}m{#1}}

\def\tsc#1{\csdef{#1}{\textsc{\lowercase{#1}}\xspace}}
\tsc{WGM}
\tsc{QE}
\tsc{EP}
\tsc{PMS}
\tsc{BEC}
\tsc{DE}
\usepackage[numbers]{natbib}
\begin{document}
\sloppy

\let\WriteBookmarks\relax
\def\floatpagepagefraction{1}
\def\textpagefraction{.001}
\shorttitle{}
\shortauthors{Jusang Lee et~al.}

\title [mode = title]{Tooth Instance Segmentation from Cone-Beam CT Images through Point-based Detection and Gaussian Disentanglement}
\tnotemark[1]

\author[1]{Jusang Lee}
\author[2]{Minyoung Chung}[orcid=0000-0001-7503-3307]\cormark[1]
\author[1]{Minkyung Lee}
\author[1]{Yeong-Gil Shin}

\address[1]{Department of Computer Science and Engineering, Seoul National University, 1 Gwanak-ro, Gwanak-gu, Seoul, 08826, Republic of Korea}
\address[2]{School of Software, Soongsil University, 369 Sangdo-Ro, Dongjak-Gu, Seoul, 06978, Republic of Korea}

\cortext[cor1]{Corresponding author (chungmy@ssu.ac.kr).}

% %%%%%%%%%% Title %%%%%%%%%%
% \title{Tooth Instance Segmentation from Cone Beam CT Scans via Point-based Detection and Gaussian Disentanglement}
% \author{Jusang Lee$^1$ \quad Minyoung Chung$^2$\thanks{The corresponding author (chungmy@ssu.ac.kr).} \quad Minkyung Lee$^1$ \quad Yeong-Gil Shin$^1$\\\\
% $^1$Seoul National University, Republic of Korea\\
% $^2$Soongsil University, Republic of Korea
% % 1 Gwanak-ro, Gwanak-gu, Seoul 08826, Korea\\
% % {\tt\small \{chungmy, ewfwa, psu9808, yshin\}@snu.ac.kr}
% }

%%%%%%%%%% Abstract %%%%%%%%%%
\begin{abstract}
    Individual tooth segmentation and identification from cone-beam computed tomography images are preoperative prerequisites for orthodontic treatments. Instance segmentation methods using convolutional neural networks have demonstrated ground-breaking results on individual tooth segmentation tasks, and are used in various medical imaging applications. While point-based detection networks achieve superior results on dental images, it is still a challenging task to distinguish adjacent teeth because of their similar topologies and proximate nature. In this study, we propose a point-based tooth localization network that effectively disentangles each individual tooth based on a Gaussian disentanglement objective function. The proposed network first performs heatmap regression accompanied by box regression for all the anatomical teeth. A novel Gaussian disentanglement penalty is employed by minimizing the sum of the pixel-wise multiplication of the heatmaps for all adjacent teeth pairs. Subsequently, individual tooth segmentation is performed by converting a pixel-wise labeling task to a distance map regression task to minimize false positives in adjacent regions of the teeth. Experimental results demonstrate that the proposed algorithm outperforms state-of-the-art approaches by increasing the average precision of detection by 9.1\%, which results in a high performance in terms of individual tooth segmentation. The primary significance of the proposed method is two-fold: 1) the introduction of a point-based tooth detection framework that does not require additional classification and 2) the design of a novel loss function that effectively separates Gaussian distributions based on heatmap responses in the point-based detection framework.
\end{abstract}

\begin{keywords}
\sep Distance-based segmentation
\sep Gaussian disentanglement loss
\sep Instance segmentation
\sep Point-based object detection
\sep Tooth CBCT segmentation
\end{keywords}

\maketitle
%\thispagestyle{empty}

%%%%%%%%%% Intro %%%%%%%%%%
\section{Introduction}
Cone-beam computed tomography (CBCT) is a widely used medical imaging technique that provides high-resolution 3D volumetric data. Recently, the development of several computer-aided diagnosis systems, along with digitalization of medical data, have helped clinicians reduce a significant amount of their workload pertaining to redundant tasks such as manual labeling and the classification of medical data \citep{van2001computer}. In particular, automating individual tooth segmentation based on CBCT images can significantly aid clinicians during orthodontic treatments in daily clinics.

Several previous studies have conducted automatic individual tooth detection and segmentation based on dental images \citep{chung2020pose, chung2020individual, cui2019toothnet}. Among these studies, convolutional neural networks (CNNs) are gaining interest because of their superior results when compared to conventional methods (e.g., level-set or statistical shape-based methods). Particularly, instance segmentation frameworks are showing promising performance for detecting target structures. An instance segmentation framework typically consists of the segmentation of individual objects that are detected and classified based on the preceding object detection process. Therefore, successful object detection is a key factor in the final performance of instance segmentation. However, accurate detection of an individual tooth is challenging, primarily because of the similar topologies of the teeth, their proximate positions, the unclear boundaries between teeth, and the presence of metal artifacts \citep{chung2020pose}.

In this study, we propose a point-based detection network using a CNN for individual tooth detection from CBCT images. We used the standard developed by the International Organization for Standardization (ISO) for anatomical tooth numbering. The point-based detection framework localizes objects of fixed classes, which indicates that it does not require any additional classification network \cite{zhou2019objects}. We used heatmap regression, which effectively preserves spatial information in images by exploiting a fully convolutional network (FCN) \citep{long2015fully}. A novel Gaussian disentanglement loss (GD loss) function is employed for training the heatmap regression. The proposed GD Loss aims to penalize the network, which prevents the output Gaussian distributions (i.e., heatmap responses) from mutual overlaps. It is designed as a non-parametric method to effectively disentangle multiple Gaussian distributions by avoiding the use of parametric measures, i.e., mean and standard deviation, which can be considered complex in the application. The size of each detected tooth is also regressed, which yields the final bounding box results of tooth detection. After the detection process, a U-Net-based segmentation network \citep{ronneberger2015u} is employed for individual tooth segmentation. To reduce false responses to adjacent teeth, we converted the pixel-wise classification task to a distance map regression task. The network regresses the distance of each pixel from the surface of the central tooth.

The remainder of this paper is organized as follows. In Section 2, we briefly review related works on object detection, as well as works related to individual tooth detection and segmentation based on dental images. The proposed method is described in Section 3, and Section 4 presents the experimental results. Finally, Section 5 presents the discussion and conclusions pertaining to this study.

%%%%%%%%%% Related Works %%%%%%%%%%
\section{Related Works}
In this section, we first review the modern CNN-based object detection methods. While most of the state-of-the-art methods were developed based on anchor-based approaches, recent studies have developed point-based methods that do not require anchor boxes. Subsequently, we briefly review related works based on tooth detection and segmentation from dental images.

\subsection{Anchor-based Object Detection}
Object detection is used to localize and classify multiple objects from images \citep{felzenszwalb2009object}. Such a process is critical for the complete understanding of an image; therefore, it is considered to be a fundamental task in computer vision \citep{zhao2019object}. Several conventional object detection methods were proposed that use handcrafted features and shallow trainable networks \citep{lowe2004distinctive, dalal2005histograms, lienhart2002extended}; however, these methods were easily outperformed by deep learning methods. Region-based CNN (R-CNN) \citep{girshick2014rich} was the first network that employed a CNN for object detection. Several studies have been conducted to improve R-CNNs, and Faster R-CNN \citep{ren2015faster} is one of the most frequently used object detection algorithms. The Faster R-CNN \citep{ren2015faster} is also called a two-stage detector, because it first produces a large number of candidate boxes using a region proposal network, and then performs classification and bounding box regression on each proposal. Other types of modern object detection algorithms can be viewed as one-stage detectors \citep{redmon2017yolo9000, redmon2016you, redmon2018yolov3}, as they directly regress the bounding box coordinates and class probabilities from image pixels.

A common aspect of both one-stage and two-stage detectors is that they both employ anchor boxes. An anchor box refers to a set of predefined boxes that are used to crop an image or feature map. The utilization of various sizes and ratios of anchor boxes helps in detecting objects that are existing in different scales in an image \cite{ren2015faster, redmon2016you}. However, anchor-based methods have certain drawbacks. First, using anchor box produces a large number of proposed regions, requiring exhaustive classification for all regions. Second, it is difficult to define optimal combinations for predefined anchors. This leads to multiple hyperparameter choices, making the designing of the network more complicated \citep{law2018cornernet}. Moreover, using an anchor box results in the production of multiple bounding boxes that overlap an object; consequently, an additional post-processing step is required. A non-maximum suppression algorithm is commonly used for removing redundant bounding boxes to obtain only one box per object, which indicates that post-processing increases the overall computational time for object detection.

\subsection{Point-based Object Detection}
Recently, object detection methods that adopt keypoint estimation have been proposed. The primary strength of point-based methods when compared to anchor-based detection is that it does not require an anchor box. Consequently, point-based methods do not suffer from the problems caused by anchor boxes (i.e., the production of a large number of proposed regions and complicated hyperparameter choices of the network). CornetNet was the first approach to introduce point-based detection \citep{law2018cornernet}, which detects objects by their top-left and bottom-right corner points. It first detects all corner points for each class, and then groups the top-left and bottom-right points into pairs for each object so that each pair of corner points results in a bounding box of the corresponding object. Vector embedding is also required for the successful grouping of corner points \citep{law2018cornernet}. Unlike CornerNet, CenterNet \citep{duan2019centernet} was proposed to directly detect the center points of objects, indicating that it does not require the training of embedding vectors for grouping. Additional regression, such as the regression of bounding box size, is performed on each object for the final detection \citep{duan2019centernet}.

\subsection{Individual Tooth Segmentation}
Several studies have been conducted on individual tooth detection and segmentation based on dental images. ToothNet \citep{cui2019toothnet} employed Mask R-CNN \citep{he2017mask}, which is an instance segmentation network based on Faster R-CNN \citep{ren2015faster}, extended it to include 3D data, and introduced a similarity matrix for the classification of different types of teeth. In \citep{chung2020pose}, the authors also employed Mask R-CNN \citep{he2017mask} as a tooth detector; howver, additional training was required to divide the upper and lower jaws to extract the volume-of-interest regions. The authors also proposed the conversion of pixel-wise classification to distance map regression to reduce false responses in the surrounding teeth \citep{chung2020pose}. In \citep{chung2020individual}, the authors dealt with dental panoramic X-ray images and localized an individual tooth using a point-based approach. While directly regressing the coordinates of all anatomical teeth, spatial distance regularization loss was employed to effectively separate the predicted coordinates.

%%%%%%%%%% Methodology %%%%%%%%%%
\section{Methodology}
%The proposed method of individual tooth detection and segmentation consist of three key factors: heatmap regression, Gaussian disentanglement, and distance map regression. The network first performs heatmap regression accompanied by bounding box size regression for individual tooth detection. A novel Gaussian Disentanglement Loss is applied at training. For each detected boxes, the U-Net based segmentation network is applied for individual tooth segmentation, which adopted distance map regression.

\subsection{Network Architecture}
\begin{figure*}[ht]
    \centering
    \includegraphics[width=0.9\textwidth]{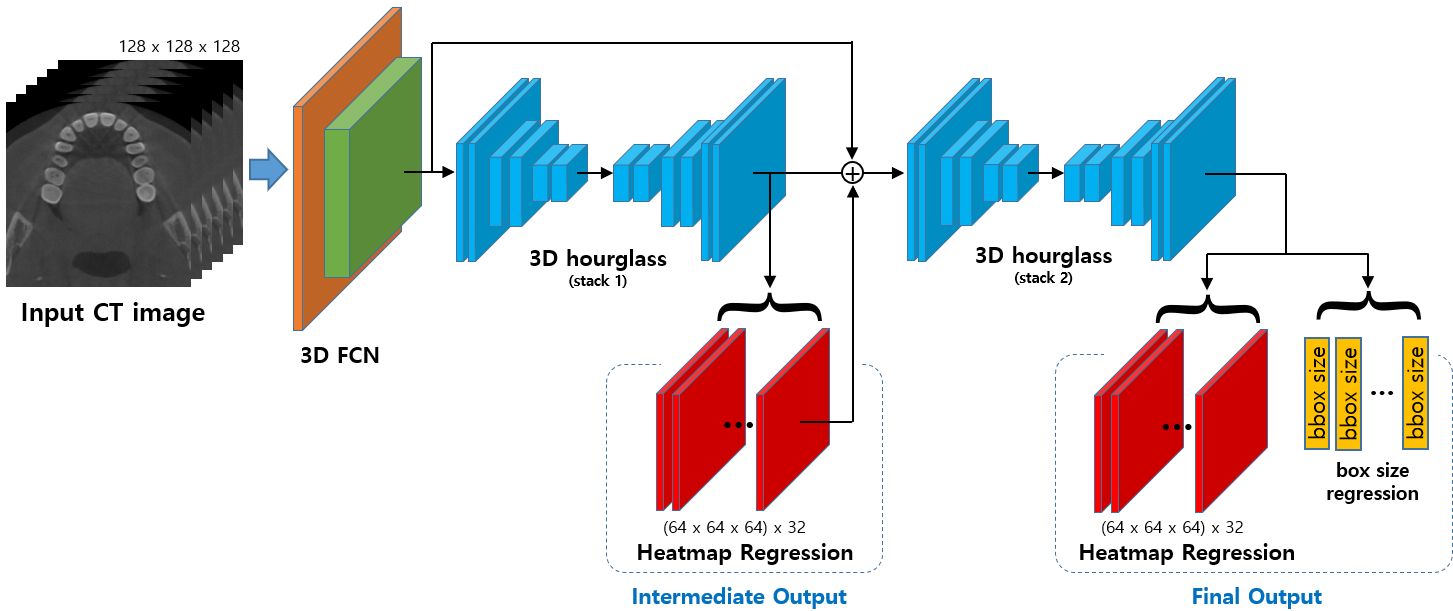}
    \caption{Overall network for tooth detection. The FCN layers followed by an encoding-decoding structure extracts feature maps from the input images. We used 3D hourglass \citep{newell2016stacked} as a backbone of the network;, and it follows two-stack architecture for an intermediate supervision. The final outputs of the network are heatmaps and bounding box sizes.}
    \label{fig:detection_network}
\end{figure*}

\begin{figure}[ht]
    \centering
    \includegraphics[width=\linewidth]{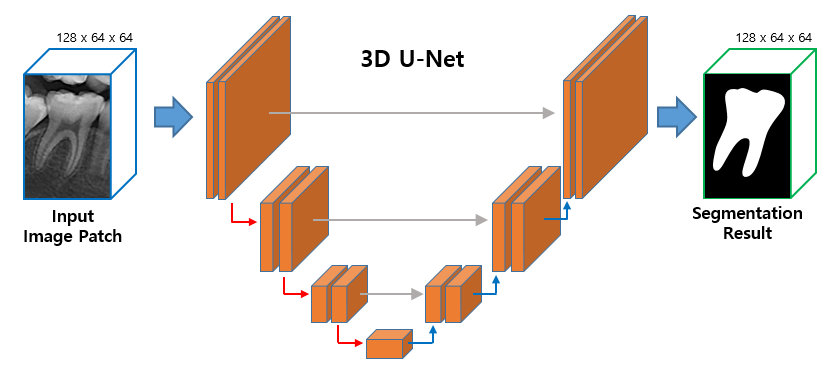}
    \caption{Overall network for tooth segmentation. U-Net architecture \citep{ronneberger2015u} is used and extended to a 3D structure. Input images are cropped into image patches by the detected bounding boxes for individual tooth segmentation.}
    \label{fig:segmentation_network}
\end{figure}

The overall architecture of the proposed network is shown in Fig. \ref{fig:detection_network}. Images of size 128 x 128 x 128 pixels were used as inputs to the network. The network extracts feature maps from input images by feeding them to the FCN layers followed by an encoding-decoding structure. We used an hourglass network \citep{newell2016stacked} as the backbone of our network, which was extended to a 3D structure. The network follows a two-stack architecture, which exploits intermediate supervision by applying the in-between heatmap results into the loss function. The output heatmaps from the first stack are then added to the previous feature maps and fed to the second stack. In the last step, the network produces the final heatmap output and bounding box sizes. After the detection process, a second network for tooth segmentation (Fig. \ref{fig:segmentation_network}) is incorporated. The input image is cropped by the detected bounding boxes into image patches and resized to 128 x 64 x 64 pixels. Subsequently, the image patches are fed to the segmentation network, i.e., U-Net, in which all convolutions are extended to 3D counterparts.


\subsection{Training the Network}
When training the detection network, three different loss functions were applied: focal loss \citep{law2018cornernet} for heatmap regression, mean-squared error (MSE) loss for bounding box size regression, and GD Loss. To train the segmentation network, we used the MSE loss instead of the binary cross-entropy (BCE) loss for distance map regression.

\subsubsection{Heatmap Regression}
As mentioned above, the proposed detection network performs a heatmap regression. As there are 32 different types of anatomical teeth, the network outputs 32 heatmaps, which are performed by 32 individual CNN layers attached parallel to each other after feature map extraction. For the loss function, we used focal loss \citep{law2018cornernet}, which primarily contributes to the imbalance of ground truth data, which is commonly used for keypoint estimation. There is a significant imbalance issue in the ground truth heatmap because it predominantly consists of a zero-valued background, in contrast to a few non-zero values in the foreground. Therefore, it is difficult to train the network using a simple MSE loss; consequently, the focal Loss is employed for heatmap regression. The focal loss can be expressed as
\begin{align}\label{eq:focal_loss}
L_{heat} & = Focal(\textbf{x}_{_{heat}}, \textbf{y}_{_{heat}})\nonumber\\
         & = \frac{-1}{N} \sum_{p}^{} 
\begin{cases}
(1-\textbf{x}(p))^{\alpha}\log{(\textbf{x}(p))}&\text{if }\textbf{y}(p)=1\\
(1-\textbf{y}(p))^{\beta}\,\textbf{x}(p)^{\alpha}\log{(1-\textbf{x}(p))}&\text{otherwise,}
\end{cases}
\end{align} where $\textbf{x}_{heat}$ is the heatmap output of the network and $\textbf{y}_{heat}$ is the heatmap of the ground truth image with pixel values of $\textbf{x}(p)$ and $\textbf{y}(p)$ at pixel p, respectively. As described in (\ref{eq:focal_loss}), the computation of the loss values is different at peak and non-peak positions. After adding the loss values from every pixel, the final loss is divided by the number of peaks. $\alpha$ and $\beta$ are the weighting coefficients that define the weights for $\textbf{x}_{heat}$ and $\textbf{y}_{heat}$ of the total loss function, respectively. The weights were set to $\alpha=2$ and $\beta=4$ to increase the impact of peak points on the total loss.

The overall network followed a two-stack architecture to employ intermediate supervision. Intermediate supervision is known to be more effective and results in better performance when compared to simply adding the same number of complexities to the network \citep{newell2016stacked}. The loss function of intermediate supervision is defined as
\begin{align}\label{eq:intermediate}
L_{heat} = \sum_{s}^{} w(s) Focal(\textbf{x}_{_{heat}}(s), \textbf{y}_{_{heat}}),
\end{align} where $\textbf{x}_{_{heat}}(s)$ is the heatmap output at each stack s and $\textbf{y}_{heat}$ is the heatmap of the ground truth image. $w(s)$ is the weighting coefficient that defines the weights for each intermediate output that affects the total loss. We set all values of $w(s)$ to one.

The network also regresses bounding box sizes to estimate the final detection boxes from each detected center point. The final regression is trained by using the MSE loss as follows:
\begin{equation}\label{eq:bbox_loss}
L_{bbox} = MSE(\textbf{x}_{bbox}, \textbf{y}_{bbox}),
\end{equation} where $\textbf{x}_{bbox}$ is the previous output of the box regression and $\textbf{y}_{bbox}$ is the ground truth. The detected center points can be regressed by localizing the peak point for each heatmap as follows:
\begin{align}\label{eq:centers}
P_{centers} = \{\underset{p}{argmax}\,(\textbf{x}_{t}(p))\,| \text{ for }t\in[1,32]\}.
\end{align}
After the center points, i.e., $P_{center}=(x_{center}, y_{center}, z_{center})$, of each tooth and the dimensions of the box $(W_{bbox}, H_{bbox}, D_{bbox})$ are regressed, the final individual tooth detection result can be expressed as follows:
\begin{align}\label{eq:bbox_final}
Bbox = (&x_{center} - W_{bbox} / 2,\,x_{center} + W_{bbox} / 2,\nonumber\\
        &y_{center} - H_{bbox} / 2,\,y_{center} + H_{bbox} / 2,\\
        &z_{center} - D_{bbox} / 2,\,z_{center} + D_{bbox} / 2).\nonumber
\end{align}

\subsection{Gaussian Disentanglement Loss}
A novel GD loss was introduced while training the heatmap regression. In dental CBCT images, adjacent teeth are positioned significantly close to each other, and they are considerably similar in shape. Thus, it is significantly difficult to train the network to accurately regress the Gaussian distributions for each tooth. The occurrence of multiple false responses on proximate regions of a tooth (i.e., severe overlapping of Gaussian distributions for two or more teeth) hinder the final performance of tooth detection.
\par

The goal of GD Loss is to ease the difficulty of achieving accurate heatmap regression by aiding the network to separate the overlapping Gaussian distributions. The proposed GD Loss presents a higher penalty when the two Gaussian distributions have greater overlaps. It is designed to avoid determining the inter-distances between Gaussian distributions, which typically involves parameters such as the mean and standard deviation. Therefore, for simplicity, we suggest a non-parametric method of separating the Gaussian distributions.

\begin{figure}[t]
    \centering
    \includegraphics[width=\linewidth]{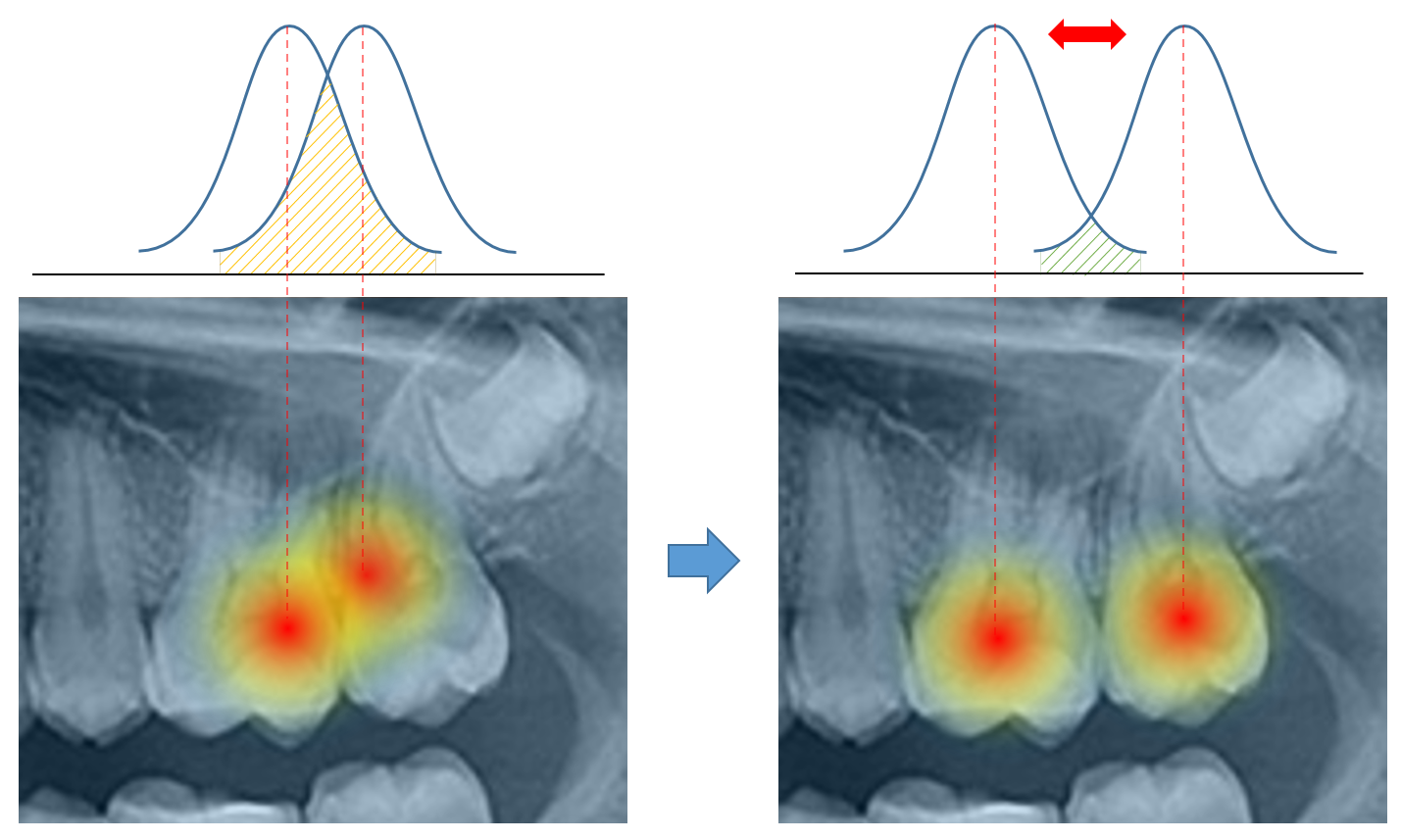}
    \caption{Non-parametric Gaussian disentanglement}
    \label{fig:nonparemetric_gd_loss}
\end{figure}

As illustrated in Fig. \ref{fig:nonparemetric_gd_loss}, if two Gaussian distributions are close to each other (i.e., entangled), the probability density functions (PDF) of both distributions have high values in the overlapping area. Conversely, if they are sufficiently disentangled, the overlapping area between the two PDFs is minimized. Thus, the joint PDF function converges to zero if the overlapping area between the distributions is sparse. Based on this aspect, we defined a non-parametric GD Loss as follows:
\begin{equation}\label{eq:gd_loss}
L_{GD} = \sum_{(i,j) \in N}^{} \sum_{p}^{} \textbf{x}_{i}(p) \odot \textbf{x}_{j}(p),
\end{equation} where $\textbf{x}_{i}(p)$ and $\textbf{x}_{j}(p)$ are the pixel values at pixel p from the heatmaps of two teeth, i.e., tooth numbers i and j, respectively. The symbol $\odot$ in (\ref{eq:gd_loss}) refers to pixel-wise multiplication. The proposed operation is performed on every pair (i, j) of adjacent teeth, and subsequently, the final GD Loss is computed by adding all pixel values of the pixel-wise multiplication results. According to the ISO standard pertaining to tooth numbers, the 32 anatomical teeth are numbered from 11 to 48.

The final objective function for training the tooth detection network, including heatmap regression and bounding box regression, can be expressed as
\begin{align}\label{eq:total_loss}
L & = \lambda _{1} L_{heat} + \lambda _{2} L_{bbox} + \lambda _{3} L_{GD},
\end{align} where $\lambda _{1},\,\lambda _{2},\,\lambda _{3}$ are the weights of each loss. We used $\lambda _{1}=0.1$, $\lambda _{2}=0.1$, $\lambda _{3}=1$ during training.

\subsection{Individual Tooth Segmentation}
For individual tooth segmentation, we initially cropped the CBCT image into image patches using the detected bounding boxes, and subsequently fed the patches to the segmentation network. However, a simple pixel-wise classification method for tooth segmentation can yield several false positives because every image patch includes partial overlaps between the adjacent teeth. In this study, similar to \citep{chung2020pose}, a distance-based segmentation method was employed for accurate tooth segmentation. The pixel-wise classification task is converted to a pixel-wise distance regression task, where the distance is defined by the shortest 3D Euclidean distance from the surface of the tooth. Accordingly, for training the distance map regression, we used MSE loss instead of BCE loss, which is defined as
\begin{equation}\label{eq:dist_loss}
L_{dist}=||\textbf{x}_{dist}-\textbf{y}_{dist}||_2^2,
% L_{dist} = MSE(\textbf{x}_{dist}, \textbf{y}_{dist}),
\end{equation} where $\textbf{x}_{dist}$ is the output of the segmentation network and $\textbf{y}_{dist}$ is the ground truth distance map. To obtain the final segmentation results, we applied a threshold to the output distance map. The overall segmentation network follows the 3D U-Net architecture \citep{cciccek20163d}, which consists of three down-sampling steps followed by three up-sampling steps.

\begin{figure}[t]
    \centering
    \includegraphics[width=\linewidth]{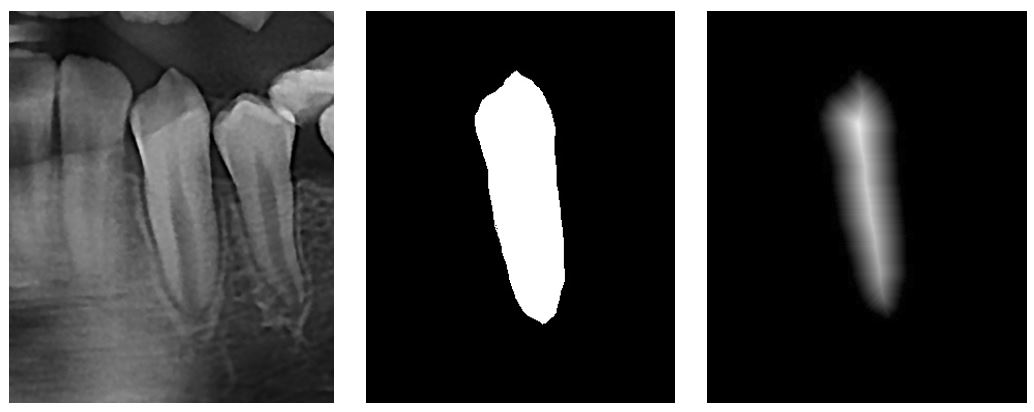}
    \caption{Two-dimensional visualization of distance map: (left) tooth image patch; (middle) ground truth mask; (right) distance map.}
    \label{fig:mask_dist_map}
\end{figure}

%%%%%%%%%% Experiments %%%%%%%%%%
\section{Experiments}
In this section, we present the evaluations of the proposed tooth detection process and the segmentation results. In total, 120 computed tomography (CT) images were collected for training, validation, and testing. We used 80 images for training, 20 images for validation, and 20 images for testing and quantitative analysis. The overall evaluation of tooth detection was conducted by comparing the proposed method with Faster R-CNN \citep{ren2015faster}, stacked hourglass network \citep{newell2016stacked}, and CenterNet \citep{duan2019centernet}, which are state-of-the-art methods for anchor-based object detection, keypoint estimation, and point-based object detection, respectively.

\subsection{Preprocessing Data}
For training the detection network, CT images were first cropped and resized to a size of 128 x 128 x 128 pixels and normalized. The ground truth data for heatmap regression were generated according to the locations of the teeth. For each anatomical tooth, a Gaussian distribution was drawn pertaining to a heatmap at the center of the tooth with a standard deviation value proportional to the tooth size.

Metadata of the tooth size for all images was also obtained and used for training. For individual tooth segmentation, CT images were cropped based on the detected bounding boxes, resized to a size of 128 x 64 x 64 pixels, and then normalized. The ground truth data for the distance map regression were subsequently generated. Each pixel inside the tooth represents the distance value from the closest surface of the tooth (i.e., unsigned value); conversely, the area outside the tooth was represented by a value of zero. We defined the distance between pixels as an \textit{approximate Euclidean distance} based on a simple two-pass algorithm \citep{rosenfeld1968distance}.

\subsection{Evaluation Metrics}
\textit{1) Intersection over Union (IOU):} The quality of detection was determined by examining the overlapping ratio between the detected bounding box and the ground truth box. IOU is used as the quantitative value of the degree of overlap. The IOU values of boxes A and B are expressed as follows:
\begin{equation}\label{eq:IOU}
IOU = \frac{|A \cap B| }{|A \cup B|}.
\end{equation}

\textit{2) Precision and Recall:} Precision and recall are frequently used for evaluating the performance of classification methods. Their values are computed as follows:
\begin{equation}\label{eq:precision}
precision = \frac{true\,positives}{true\,positives + false\,positives}
\end{equation}
and
\begin{equation}\label{eq:recall}
recall = \frac{true\,positives}{true\,positives + false\,negatives}.
\end{equation}
Precision and recall are also used for object detection. The detected bounding boxes are classified as true positives by considering the IOU values of the boxes. For example, if the IOU value is higher than a certain threshold, the corresponding detection box is regarded as a true positive sample. Conversely, the detection box is counted as a false positive sample if the IOU value is less than a given threshold.

\textit{3) Average Precision (AP):} The general definition of AP is the area under the precision-recall (PR) curve. A larger area under the PR curve indicates that the overall detection performance is high. When determining the precision and recall, we used an IOU threshold value of 50; consequently, the resulting AP was marked as AP50.

\textit{4) Object Include Ratio (OIR):} As mentioned in the previous section, accurate object detection is critical for the final performance of instance segmentation. Specifically, it is important that the detected bounding boxes should completely include the target teeth to acquire intact segmentation results. Therefore, evaluating the overlapping ratio with not only the ground truth boxes but also the actual object area is important. Such an evaluation is performed using OIR, which can be calculated as follows \citep{chung2020pose}:
\begin{equation}\label{eq:OIR}
OIR = \frac{|M \cap D|}{|M|},
\end{equation} where M is the area of the object and D is the area of the detected bounding box.

\textit{5) Confusion Matrix:} To evaluate the identification performance (i.e., the capability of distinguishing all anatomical tooth types), we present a confusion matrix regarding the classes of each tooth. Each row of the matrix represents the actual class of objects and each column of the matrix represents the predicted class. After adding all the cells of the matrix according to the prediction results, each value of the cell was normalized by the number of ground truths per class. An accurate classification result is expected to yield a confusion matrix with peak values in its diagonal elements. For greater legibility, we applied colors to the cells based on their values.

\subsection{Accuracy Evaluation}
\begin{figure*}[ht]
    \centering
    \includegraphics[width=\textwidth]{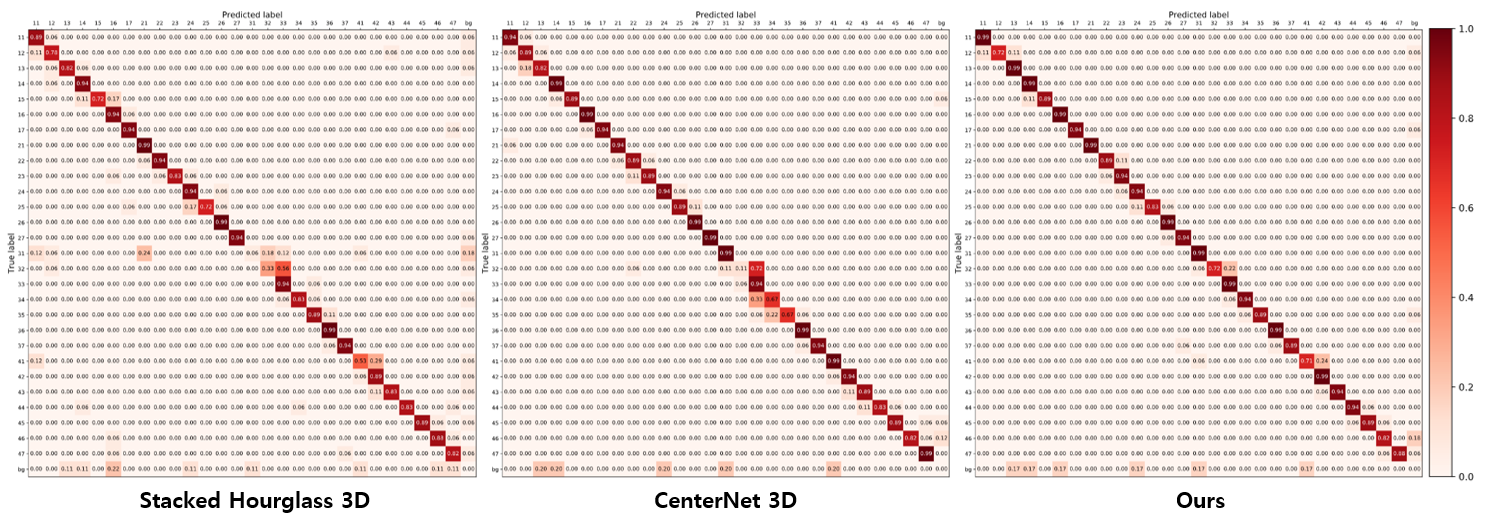}
    \caption[Confusion matrix of tooth identification]{Confusion matrix of tooth identification}
    \label{fig:cm_result}
\end{figure*}

The experiment was conducted by comparing the proposed method with state-of-the-art detection networks, including Faster R-CNN \citep{ren2015faster}, stacked hourglass network \citep{newell2016stacked}, and CenterNet \citep{duan2019centernet}. Faster R-CNN is one of the most frequently used object detectors, which is used in several applications. The stacked hourglass network \citep{newell2016stacked} is one of the most popular baselines for keypoint estimation tasks, such as human pose estimation. We included CenterNet in our comparative analysis to verify the effect of GD Loss, as the overall architecture of our proposed network is closely related to CenterNet. In all the experiments, all convolution operations of the networks were extended to the 3D counterparts to cope with the CBCT images. The accuracy of tooth detection was evaluated based on different metrics including AP50, OIR, and mean IOU, as defined previously. The accuracy of identification was also evaluated, and the results were plotted as a confusion matrix (Fig. \ref{fig:cm_result}).

\begin{table}[t]
\centering
\begin{tabular}{ m{3.5cm} | m{1cm} | m{1cm} | m{1cm}} 
    Methods & AP50 & OIR & mIOU \\ 
    \hline
    Faster R-CNN 3D         & 69.08 & 0.781 & 0.545 \\ 
    Stacked Hourglass 3D    & 72.73 & 0.822 & 0.587 \\
    CenterNet 3D            & 81.82 & 0.847 & 0.695 \\ 
    \textbf{Ours}           & \textbf{90.91} & \textbf{0.966} & \textbf{0.704} \\ 
    \textbf{Ours w/ margin} & N/A            & \textbf{0.999} & N/A \\ 
    \hline
\end{tabular}
\caption{\label{tab:det_results}Experimental results of tooth detection}
\end{table}

\begin{table}[t]
\centering
\begin{tabular}{ m{3.5cm} | m{1.5cm} | m{1.5cm}} 
    Methods & Precision & Recall \\ 
    \hline
    Faster R-CNN 3D         & 0.806 & 0.731 \\ 
    Stacked Hourglass 3D    & 0.853 & 0.826 \\
    CenterNet 3D            & 0.891 & 0.883 \\ 
    \textbf{Ours}           & \textbf{0.932} & \textbf{0.919}\\ 
    \hline
\end{tabular}
\caption{\label{tab:cls_results}Experimental results of tooth identification}
\end{table}

The performance evaluation results of the tooth detection process are listed in Table \ref{tab:det_results}. The table demonstrates that our proposed method is superior to other state-of-the-art methods. Moreover, the addition of margins of 5 to 10 pixels at the detected tooth bounding boxes resulted in achieving an OIR value of 0.999. This indicates that the bounding boxes with margins almost completely included the tooth objects, which is critical for the subsequent segmentation process.

The evaluation results of the tooth identification process are listed in Table \ref{tab:cls_results}. The results show that our proposed method performs better than the other methods in the tooth identification task. Each detected box is counted as a true positive if the IOU value is higher than the threshold (i.e., 50), and false positive if not. The precision and recall values were computed based on these counts. The identification result was plotted as a confusion matrix (Fig. \ref{fig:cm_result}) for visual representation. While other networks failed to identify adjacent teeth, especially for small teeth (tooth numbers 31, 32, 41, and 42), our proposed method demonstrated better quality of identification, as indicated by the diagonal elements of the confusion matrix, which had the highest values.

\subsection{Gaussian Disentanglement}
\begin{figure*}[ht!]
    \centering
    \includegraphics[width=\textwidth]{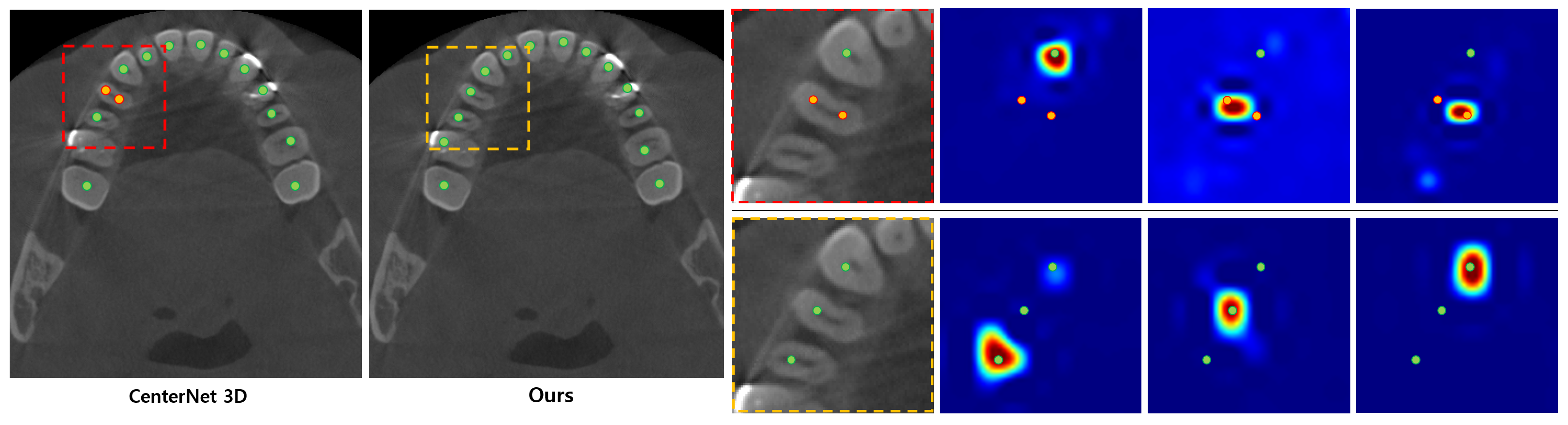}
    \caption[Effect of Gaussian Disentanglement]{Effect of Gaussian disentanglement loss}
    \label{fig:gd_vis}
\end{figure*}

As shown in the experimental results, distinguishing adjacent teeth is the primary challenge in achieving high performance in the tooth detection and identification processes. Based on the basic heatmap regression, the Gaussian distributions of adjacent teeth tend to overlap at the proximate position. We successfully addressed the overlapping problem in our method by introducing GD loss. As illustrated in Fig. \ref{fig:gd_vis}, the heatmap regression by CenterNet \citep{duan2019centernet} showed overlapping Gaussian distributions, resulting in the false detection of the tooth center. Conversely, our method showed heatmaps with Gaussian distributions that are successfully disentangled from each other.

\begin{figure*}[ht!]
    \centering
    \includegraphics[width=\textwidth]{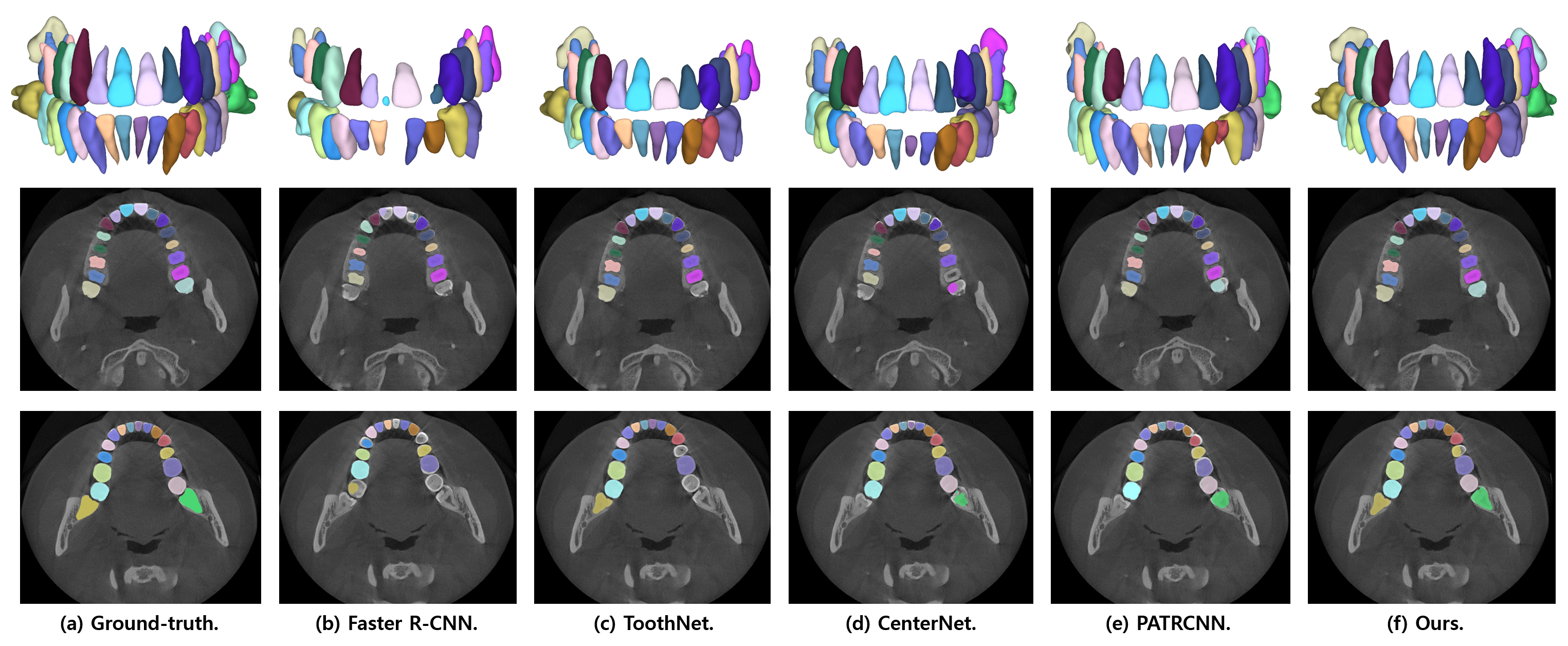}
    \caption[Result of individual tooth segmentation]{Visualization of the segmentation results on the test image. The columns indicate (a) Ground truth label, (b) Faster R-CNN \citep{ren2015faster}, (c) ToothNet \citep{cui2019toothnet}, (d) CenterNet \citep{duan2019centernet}, (e) pose-aware tooth R-CNN \citep{chung2020pose}, and (f) our method. Note that Faster R-CNN, CenterNet, and our method performed individual tooth detection, followed by tooth segmentation using 3D U-Net \citep{cciccek20163d}.}
    \label{fig:seg_result}
\end{figure*}

\subsection{Tooth Segmentation}
%As mentioned above, adding small margins to detected bounding boxes of our method resulted in OIR value of 0.999, which means that the detected boxes fully include the tooth areas. As shown in Fig. \ref{fig:seg_result}, such improvement resulted in accurate segmentation of individual tooth in CBCT images.
The segmentation results of the proposed method are illustrated in Fig. \ref{fig:seg_result}. The results were compared with other state-of-the-art individual tooth detection and segmentation methods, i.e., Faster R-CNN \citep{ren2015faster}, ToothNet \citep{cui2019toothnet}, CenterNet \citep{duan2019centernet}, and pose-aware tooth R-CNN \citep{chung2020pose}. Similar to our proposed method, Faster R-CNN and CenterNet first performed tooth detection, followed by tooth segmentation by employing 3D U-Net \citep{cciccek20163d}. According to Fig. \ref{fig:seg_result}, the segmentation result of our method shows superior performance when compared to the other methods. Faster R-CNN and CenterNet failed to segment several teeth, predominantly the frontal teeth. Moreover, it can be observed that missing a single tooth resulted in successive misidentification of several other adjacent teeth. In particular, in the second row of Fig. \ref{fig:seg_result}d, the first left premolar (tooth number 24) was misidentified as the second left premolar (tooth number 25), which led to the misidentification of the second left premolar, first left molar, and second left molar, causing the left wisdom tooth to be undetected. ToothNet \citep{cui2019toothnet} was successful in tooth identification, but it failed to segment certain teeth. PATRCNN \citep{chung2020pose} showed comparable performance to our method, but our method outperformed PATRCNN with more accurate tooth detection.

%%%%%%%%%% Discussion + Conclusion %%%%%%%%%%
\section{Discussion and Conclusion}
In this study, we proposed a point-based detection network for accurate tooth detection and segmentation from CBCT images. Detecting an individual tooth from CBCT images is considered a challenging task because of the similar topologies of teeth and their proximate positions. We proposed an effective point-based detection method that outperforms general object detection methods \citep{zhao2019object}. By employing a point-based detection framework \citep{law2018cornernet, duan2019centernet}, we identified different types of anatomical teeth without training any additional classification module, which is a limitation of the anchor-based detectors \citep{girshick2014rich, ren2015faster, chung2020pose}. Moreover, the proposed method accurately detected adjacent teeth by introducing a novel GD loss function within heatmap regression. Improvements in the performance of tooth detection resulted in accurate individual tooth segmentation.

Further improvements may include identifying missing teeth. Because our point-based tooth detector detects all anatomical teeth, it attempts to detect teeth that are not present in the input CBCT image. Although the proposed method has successfully performed tooth detection and identification, an additional light-weight classification process might be required to determine the presence of absence of teeth. Furthermore, because the teeth in CBCT images show various angles, regressing their poses and realigning them will improve data consistency, which will lead to improved segmentation performance.

\section*{Acknowledgement}
This work was supported by HealthHub Inc. (No.0536-20200006, Development of Artificial Disk Simulation Software for Cervical Vertebrae Surgery).

%%%%%%%%%% References %%%%%%%%%%
\bibliographystyle{bibformat}
\bibliography{mybib}

\vskip3pt
\bio{}
\textbf{Jusang Lee} is a M.S. course student in the Department of Computer Science and Engineering, Seoul National University, South Korea. His research interests include medical image processing, computer vision, and machine intelligence (AI; deep learning).
\endbio

\bio{}
\textbf{Minyoung Chung} is an assistant professor in School of Software at Soongsil University, South Korea. He received the Ph.D. degree (2020) in Computer Science and Engineering at Seoul National University, South Korea. His current research interests include medical image processing, computer vision, machine intelligence (AI; deep learning), 3D visualization, GPU parallel processing, imaging software development, and human-computer interactions.
\endbio

\bio{}
\textbf{Minkyung Lee} is a Ph.D. course student in the Department of Computer Science and Engineering, Seoul National University, South Korea. Her research interests include medical image processing, computer vision, and machine intelligence (AI; deep learning).
\endbio

\bio{}
\textbf{Yeong-Gil Shin} is a professor in the Department of Computer Science and Engineering and the Director of Computer Graphics and Image Processing Laboratory, Seoul National University, South Korea. He received the B.S. (1981) and M.S. (1984) degrees in Computer Science and Engineering from Seoul National University, South Korea. He received the Ph.D. degree (1990) in Computer Science at the University of Southern California. His current interests are medical image processing, deep learning, and real-time 3D volume rendering.
\endbio

\end{document}